\documentclass[runningheads]{llncs}
\usepackage[T1]{fontenc}
\newcommand{\parhead}[1]{\noindent\textbf{#1}\,}
\usepackage{graphicx,verbatim}
\usepackage[numbers,sort&compress]{natbib}
\usepackage[pagebackref, breaklinks=true, colorlinks, citecolor=citecolor, bookmarks=false]{hyperref}

\usepackage{amsmath,amssymb,amsfonts}
\usepackage{algorithmic}
\usepackage{algorithm}
\usepackage{graphicx}
\usepackage{multirow}
\usepackage{xcolor}
\usepackage{nicematrix}
\usepackage{float}
\usepackage{upgreek}
\definecolor{lightgreen}{RGB}{231,245,234}
\definecolor{lightpurple}{RGB}{236,223,247}
\definecolor{lightred}{RGB}{255,204,153}
\definecolor{tabhighlight}{HTML}{e5e5e5}
\usepackage{bm}
\usepackage{colortbl}
\usepackage{tablefootnote}
\usepackage{booktabs}
\usepackage{pifont}
\definecolor{tabhighlight}{HTML}{e5e5e5}
\definecolor{citecolor}{rgb}{0.21,0.49,0.74}
\definecolor{mutedblue}{RGB}{70, 90, 120}
\usepackage{amsmath,amssymb}
\usepackage{wrapfig}
\usepackage{multirow}
\usepackage{marvosym}
\usepackage{hyperref}
\usepackage{bm}
\usepackage{graphicx}
\usepackage{colortbl}
\usepackage{booktabs}
\renewcommand{\thefootnote}{\fnsymbol{footnote}}
\usepackage{makecell}
\usepackage{pifont}
\usepackage{caption}

\newcommand{\tablestyle}[2]{\setlength{\tabcolsep}{#1}\renewcommand{\arraystretch}{#2}\centering\footnotesize}
\def\BibTeX{{\rm B\kern-.05em{\sc i\kern-.025em b}\kern-.08em
    T\kern-.1667em\lower.7ex\hbox{E}\kern-.125emX}}

\usepackage{color}

\begin{document}
\title{Evi-Steer: Learning to Steer Biomedical Vision-Language Models through Efficient and Generalizable Evidential Tuning}
\titlerunning{Biomedical Multi-modal Low-dimensional Evidential Steering}

\author{Taha Koleilat\textsuperscript{\Letter} \and
Hassan Rivaz\and
Yiming Xiao}
\institute{Concordia University, Montreal, Canada}


\authorrunning{T. Koleilat et al.}

\maketitle              

\renewcommand{\thefootnote}{}
\footnote{\Letter\ Corresponding Author: \href{mailto:taha.koleilat@mail.concordia.ca}{taha.koleilat@mail.concordia.ca}}

\begin{abstract}
Parameter-efficient adaptation of vision--language foundation models is crucial for precise multimodal understanding of biomedical images, yet existing methods remain deterministic and often struggle under domain shift or ambiguous image--text alignment. This limitation is particularly critical in the clinic, where models should remain robust in low-data regimes and domain shifts. We present \texttt{Evi-Steer}, an \emph{evidential cross-modal low-dimensional steering} framework for BiomedCLIP that enables uncertainty-aware parameter-efficient fine-tuning while updating only \textbf{0.11\% of total model parameters}. Our approach performs lightweight low-dimensional token updates in both vision and text encoders while simultaneously estimating epistemic uncertainty. These uncertainty estimates update gate residuals, allowing the model to adapt conservatively when evidence is weak. Furthermore, we introduce cross-modal confidence fusion based on \textit{Dempster--Shafer theory}, enabling visual adaptation to be conditioned on textual confidence and suppressing conflicting or uncertain cross-modal updates. We conduct a comprehensive evaluation on 15 biomedical imaging datasets spanning 8 organs and 8 imaging modalities under few-shot learning and domain generalization settings. \texttt{Evi-Steer} consistently outperforms state-of-the-art methods under few-shot learning and domain shift settings, demonstrating a practical and robust pathway for deploying vision-language models in real-world clinical settings. Code is available at \url{https://github.com/HealthX-Lab/Evi-Steer}.


\end{abstract}

\section{Introduction}

Vision–language foundation models (VLMs), such as OpenAI’s CLIP \cite{radford2021learning} and domain-specific variants such as BiomedCLIP \cite{biomedclip}, provide a promising route to label-efficient learning by aligning images and natural language through large-scale contrastive pretraining. This paradigm is particularly attractive in biomedical settings \cite{koleilat2024medclip,koleilat2024medclipv2}, where expert-labeled data are scarce and expensive, images exhibit subtle inter-class differences and high intra-class variability across scanners and patient populations, and domain shifts often degrade out-of-distribution (OOD) generalization. In addition to that, practical clinical deployment demands resource- and parameter-efficient adaptation for real-time use. However, effective biomedical adaptation requires more than independent tuning of vision and text encoders. Clinical semantics are inherently cross-modal, and robust transfer demands coordinated co-adaptation of visual and textual representations. Simply updating one modality or injecting static prompts may fail to resolve cross-modal inconsistencies \cite{spiegler2025textsam, koleilat2026medclipseg, rasaee2025grounding}, especially when prompts are ambiguous or visual features are subtle.

Full fine-tuning of large VLMs is computationally expensive and can degrade pretrained knowledge \cite{shuttleworth2025lora}, especially in few-shot settings. This has motivated parameter-efficient adaptation. Low-rank adaptation (LoRA) \cite{hu2021lora,manzari2026sparse,zanella2024low} and related methods \cite{sun2022singular,koleilat2025singular} operate in the model weight space by learning low-rank updates to internal projection matrices. Prompt-based approaches \cite{zhou2022learning,zhou2022conditional,khattak2023maple} modify input representations by introducing trainable tokens, while adapter-based methods \cite{zhang2021tip,gao2024clip} insert lightweight bottleneck modules that transform intermediate activations. Although these approaches reduce the number of trainable parameters, they remain deterministic, applying residual updates without accounting for uncertainty in the adaptation process. So far, most uncertainty-aware deep learning methods rely on sampling-based Bayesian inference~\cite{gal2016dropout,kendall2017uncertainties}, which requires multiple forward passes and incurs substantial overhead. Evidential inference instead estimates evidence and uncertainty in a single pass~\cite{amini2020deep,sensoy2018evidential}, making it attractive for efficient VLM adaptation. Recent works have explored related directions: CILMP applies ReFT-style low-rank interventions for disease-specific prompting~\cite{du2025medical}, while EviVLM and Bayesian-PEFT incorporate evidential uncertainty into segmentation and PEFT, respectively~\cite{pan2025evivlm,pandey2024be}. However, they do not jointly perform cross-modal evidential steering in activation space to enable efficient adaptation while preserving generalizable pretrained representations.

Motivated by this gap, we present \texttt{Evi-Steer}, an \emph{evidential cross-modal low-dimensional steering} module for parameter-efficient tuning of BiomedCLIP. In contrast to encoder weight-space adaptation methods, we follow a \emph{representation steering} paradigm \cite{wu2024reft}, which performs adaptation directly in the activation space while keeping the backbone parameters frozen. Specifically, we inject lightweight $r$-dimensional cross-modal modules into intermediate layers of both the vision and text encoders to modulate token representations rather than altering pretrained weights. To achieve uncertainty-aware adaptation, we estimate intermediate confidence through evidential modeling and use these estimates to gate residual updates instead of applying uniform deterministic injections. We also fuse vision and text evidence via \textit{Dempster--Shafer belief combination} \cite{shafer1992dempster} to suppress adaptation from conflicting or uncertain text-image samples. By jointly co-adapting both modalities under uncertainty gating, \texttt{Evi-Steer} preserves the pretrained knowledge of the foundation model while improving few-shot accuracy and robustness under domain shift. \underline{Our contributions are three-fold}: \textbf{First}, we introduce an \textbf{evidential representation steering} mechanism that estimates \emph{latent dimension} (LD) epistemic uncertainty and applies confidence-weighted updates directly in the activation space rather than the weight space.
\textbf{Second}, we propose \textbf{cross-modal belief fusion} that conditions visual adaptation on textual confidence, improving robustness when prompts or features are ambiguous. \textbf{Third}, we provide \textbf{comprehensive evaluation} on 15 biomedical datasets under \emph{few-shot learning} and \emph{domain generalization}, demonstrating consistent gains over SOTA baselines.

\begin{figure}[t]
    \centering
    \includegraphics[width=\linewidth]{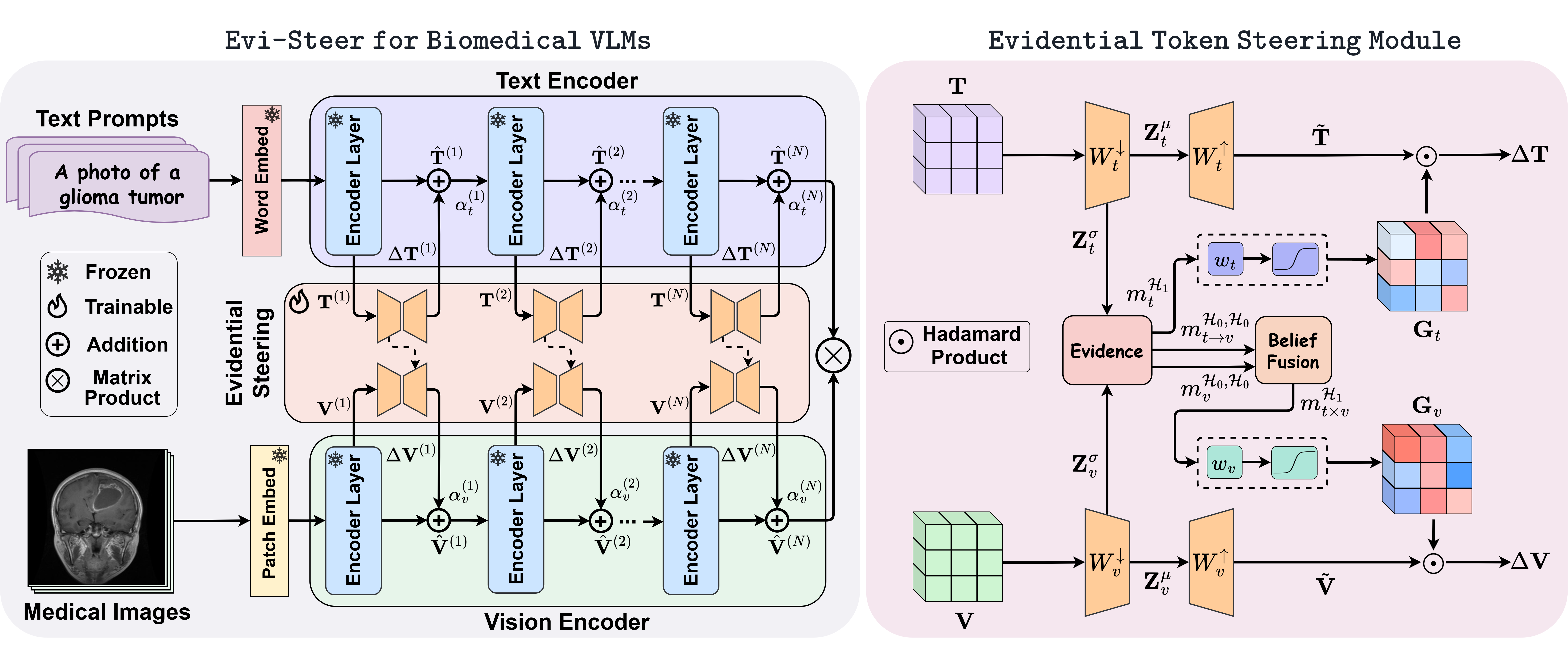}
    \caption{Overview of our framework where evidential cross-modal low-dimensional adapters produce confidence-weighted representation updates.}
    \label{fig:framework}
\end{figure}
\section{Methods and Materials}
\subsection{Model architecture of \texttt{Evi-Steer}}
\noindent Figure \ref{fig:framework} illustrates the overall pipeline of \texttt{Evi-Steer}. Given an input image and a set of class prompts, the vision and text encoders first extract pretrained features. We then inject cross-modal low-dimensional representation adapters into the first $d$ layers of both encoders. At each adapted layer, the module (i) computes a low-dimensional residual update in the activation space, (ii) estimates \emph{latent dimension} (LD) epistemic uncertainty through evidential modeling, and (iii) reweights the representation token steering based on the estimated confidence at the layer. Textual uncertainties are aggregated and converted into belief masses that are fused with visual evidence via \textit{Dempster–Shafer combination}. The fused belief modulates visual adaptation, ensuring that updates are emphasized only when both modalities provide reliable evidence. The final adapted representations are then used for classification.

\medskip
\parhead{Token representations.}
At encoder layer $n\in\{1,\dots,N\}$, let the vision tokens be
$\mathbf{V}^{(n)} \in \mathbb{R}^{B \times P_v \times D_v}$
and the text tokens be
$\mathbf{T}^{(n)} \in \mathbb{R}^{C \times P_t \times D_t}$,
where $B$ is the batch size, $\{P_v, P_t\}$ are the numbers of vision and text tokens respectively, $C$ is the number of classes, and $\{D_v, D_t\}$ are the channel dimensions of each modality. We inject cross-modal low-dimensional adapters at each of the first $d$ layers of both encoders, while keeping the backbone parameters frozen.

\smallskip
\parhead{Latent Dimension Representation.}
Unlike LoRA \cite{hu2021lora}, which approximates weight updates through low-rank decompositions of model parameters, our approach follows ReFT \cite{wu2024reft}, which directly decomposes activations to model low-dimensional \emph{steering directions} in token representations. At Layer $n$, text and vision representations are projected into a shared $r$-dimensional latent space using a single down-projection per modality. The resulting latent representation is split into mean-update and uncertainty
channels:
\begin{equation}
[\mathbf{Z}^{\mu}_t,\,\mathbf{Z}^{\sigma}_t]
=
\mathbf{T}^{(n)}W^{\downarrow}_t,
\quad
W^{\downarrow}_t\in\mathbb{R}^{D_t\times 2r}
\end{equation}
\begin{equation}
[\mathbf{Z}^{\mu}_v,\,\mathbf{Z}^{\sigma}_v]
=
\mathbf{V}^{(n)}W^{\downarrow}_v,
\quad
W^{\downarrow}_v\in\mathbb{R}^{D_v\times 2r}.
\end{equation}
where
$\mathbf{Z}^{\mu}_t,\mathbf{Z}^{\sigma}_t \in \mathbb{R}^{C \times P_t \times r}$ and
$\mathbf{Z}^{\mu}_v,\mathbf{Z}^{\sigma}_v \in \mathbb{R}^{B \times P_v \times r}$
denote the LD mean update and standard deviation representations for text and vision, respectively. The low-dimensional token updates are then obtained via their up-projections:
\begin{equation}
\tilde{\mathbf{T}}^{(n)} = \mathbf{Z}^{\mu}_t W^{\uparrow}_t,
\qquad
W^{\uparrow}_t \in \mathbb{R}^{r \times D_t},
\end{equation}
\begin{equation}
\tilde{\mathbf{V}}^{(n)} = \mathbf{Z}^{\mu}_v W^{\uparrow}_v,
\qquad
W^{\uparrow}_v \in \mathbb{R}^{r \times D_v}.
\end{equation}

\noindent As in standard adapter design, the up-projection matrices
$W^{\uparrow}_t$ (text) and $W^{\uparrow}_v$ (vision) are initialized to zero, ensuring stable warm-start training and preserving the pretrained backbone at initialization.

\smallskip
\parhead{Evidential uncertainty in latent space.}
We model LD epistemic uncertainty using an evidential construction.
For each token, the uncertainty channel $\mathbf{Z}^{\sigma}$ is mapped to evidence $\mathbf{e}$,
Dirichlet concentration parameters $\boldsymbol{\beta}$, and the corresponding LD epistemic uncertainty $\mathbf{u}$ as
\begin{equation}
\mathbf{e} = \texttt{softplus}\big((\mathbf{Z}^{\sigma})^{\circ 2}\big) + \epsilon,
\qquad
\boldsymbol{\beta} = \mathbf{e} + 1,
\qquad
\mathbf{u} = \boldsymbol{\beta}^{-1}.
\end{equation}
Here, $(\cdot)^{\circ 2}$ denotes element-wise square and $\epsilon$ is a small constant. Since latent features are not simplex-normalized class probabilities, we use $\boldsymbol{\beta}^{-1}$ as an element-wise inverse-evidence proxy for latent uncertainty in comparison to~\cite{sensoy2018evidential}.

\smallskip
\parhead{Confidence modeling.}
Each modality produces LD uncertainty estimates that quantify the
confidence of applying low-dimensional updates.
We interpret uncertainty through two complementary hypotheses:
\emph{apply update}, denoted by $\mathcal{H}_1$, and \emph{ignorance}, denoted by $\mathcal{H}_0$.
For a given uncertainty tensor $\mathbf{u}_v$/$\mathbf{u}_t$, the corresponding belief masses for vision and text are defined as
\begin{equation}
m^{\mathcal{H}_1}_{v} = 1-\mathbf{u}_v,
\qquad
m^{\mathcal{H}_0}_{v} = \mathbf{u}_v,
\end{equation}
\begin{equation}
m^{\mathcal{H}_1}_{t} = 1-\mathbf{u}_t,
\qquad
m^{\mathcal{H}_0}_{t} = \mathbf{u}_t,
\end{equation}
such that confident tokens contribute more strongly to adaptation, while uncertain components are conservatively suppressed.

\smallskip
\parhead{Cross-modal belief fusion.}
\textit{Dempster–Shafer theory} (DST) provides a principled framework for combining evidence from multiple sources while explicitly modeling uncertainty as belief and ignorance. Unlike standard probabilistic fusion, DST separates confidence from uncertainty and employs a combination rule that naturally downweights conflicting evidence. This property makes it particularly suitable for cross-modal scenarios, where visual and textual signals may be complementary, unreliable, or ambiguous. Here, we use DST to condition visual adaptation on textual confidence. Specifically, we first aggregate textual uncertainty across class prompts using the \texttt{[EOS]}\footnote{CLIP uses the \texttt{[EOS]} token as the \texttt{[CLS]} token} token to obtain an LD-based uncertainty summary. This pooled uncertainty is then converted into belief masses that modulate visual updates to encourage vision encoder adaptation when textual evidence is confident and suppress it otherwise.
\begin{equation}
\bar{\mathbf{u}}_t
=
\frac{1}{C}
\sum_{c=1}^{C}
\mathbf{u}_{t,c}^{\texttt{[EOS]}}
\in
\mathbb{R}^{1\times1\times r},
\qquad
m^{\mathcal{H}_1}_{t \rightarrow v} = 1-\bar{\mathbf{u}}_t,
\qquad
m^{\mathcal{H}_0}_{t \rightarrow v} = \bar{\mathbf{u}}_t.
\end{equation}
\noindent
The two belief sources are fused using the \textit{Dempster--Shafer combination}:
\begin{equation}
m^{\mathcal{H}_1}_{t\times v}
=
\frac{
m^{\mathcal{H}_1}_{t \rightarrow v} \, m^{\mathcal{H}_1}_{v}
}{
1 -
(m^{\mathcal{H}_1}_{t \rightarrow v} \, m^{\mathcal{H}_0}_{v}
+
m^{\mathcal{H}_0}_{t \rightarrow v} \, m^{\mathcal{H}_1}_{v})
+ \epsilon
}.
\end{equation}
The fused belief
$m^{\mathcal{H}_1}_{t\times v} \in \mathbb{R}^{B \times P_v \times r}$
captures \emph{consensus confidence} between modalities, emphasizing updates only when both text and vision provide evidence, and suppressing adaptation under ambiguity or conflict. $\epsilon$ is a small constant.

\smallskip
\parhead{Text and vision confidence.}
Linear layers $\{w_v, w_t\}$ aggregate LD belief masses along the LD axis to produce token-wise \emph{confidence weights} $\mathbf{G}$ and weighted updates ($\Updelta\mathbf{T}, \Updelta\mathbf{V}$). For text prompts, unimodal belief is used directly:
\begin{equation}
 \Updelta\mathbf{T}^{(n)} =  \mathbf{G}_t^{(n)} \odot \tilde{\mathbf{T}}^{(n)}, \quad \mathbf{G}_t^{(n)} = \texttt{sigmoid}\!\left(m^{\mathcal{H}_1}_{t} \mathbf{w}_t\right),
\quad
\mathbf{w}_t \in \mathbb{R}^{r \times 1}
\end{equation}
while for vision tokens, confidence is computed from the fused cross-modal belief:
\begin{equation}
\Updelta \mathbf{V}^{(n)} = \mathbf{G}_v^{(n)} \odot \tilde{\mathbf{V}}^{(n)}, \quad \mathbf{G}_v^{(n)} = \texttt{sigmoid}\!\left(m^{\mathcal{H}_1}_{t \times v} \mathbf{w}_v\right),
\quad
\mathbf{w}_v \in \mathbb{R}^{r \times 1}
\end{equation}
This formulation ensures consistent treatment of uncertainty across modalities, while allowing vision updates to be modulated by joint text--image confidence.

\smallskip
\parhead{Evidential token steering.}
The confidence-weighted updates are applied with learnable scalings $\{\alpha_v^{(n)}, \alpha_t^{(n)}\} \in [0,1]$ 
to control overall steering magnitude:
\begin{align}
\hat{\mathbf{V}}^{(n)} &=
\mathbf{V}^{(n)} +
\alpha_v^{(n)} * \Updelta \mathbf{V}^{(n)},\\
\hat{\mathbf{T}}^{(n)} &=
\mathbf{T}^{(n)} +
\alpha_t^{(n)} * \Updelta \mathbf{T}^{(n)}.
\end{align}

\smallskip
\parhead{Evidential regularization.}
To prevent degenerate uncertainty estimates and enforce statistically meaningful evidence, we regularize the concentration parameters $\boldsymbol{\beta}$ produced by the latent uncertainty channel by penalizing their deviation from a neutral prior $\text{Gamma}(1,1)$ via a Kullback--Leibler divergence:
\begin{equation}
\mathcal{L}_{\text{KL}}
=
KL\!\left(
\text{Gamma}(\boldsymbol{\beta},1)
\,\Vert\,
\text{Gamma}(1,1)
\right)
=
(\boldsymbol{\beta}-1)\psi(\boldsymbol{\beta})
-
\log\Gamma(\boldsymbol{\beta}),
\end{equation}
where $\psi(\cdot)$ denotes the digamma function and $\Gamma(\cdot)$ the Gamma function. The final objective combines the classification loss with this regularization:
\begin{equation}
\mathcal{L}
=
\mathcal{L}_{\text{CE}}
+
\lambda_{\text{KL}}\mathcal{L}_{\text{KL}}, \quad \lambda_{\text{KL}} = 10^{-4}
\end{equation}

\subsection{Experimental Setup}
We evaluate the proposed method on a broad collection of biomedical imaging benchmarks under protocols designed to measure both data efficiency and robustness across few-shot classification scenarios.

\noindent \textbf{Few-Shot Learning.}  
To examine performance in low-data regimes, we conduct experiments with $K=\{4,8,16\}$ labeled samples per class. This setting reflects realistic clinical scenarios where expert annotations are scarce.

\noindent \textbf{Domain Generalization.}  
To study robustness to distribution shift, models are trained using the 16-shot setting on a source dataset and evaluated on both the source and several \underline{unseen target datasets} without any additional fine-tuning. This protocol measures the ability to transfer knowledge across domains.

\noindent \textbf{Datasets.}  
Experiments are performed on 15 medical imaging datasets spanning 8 organs and 8 imaging modalities. For \emph{few-shot learning}, we use computerized tomography (CTKidney \cite{ctkidney}), endoscopy (Kvasir \cite{kvasir}), fundus photography (RETINA \cite{retina1,retina2}), histopathology (LC25000 \cite{LC25000}, CHMNIST \cite{chmnist}), magnetic resonance imaging (BTMRI \cite{btmri}), optical coherence tomography (OCTMNIST \cite{octmnist}), ultrasound (BUSI \cite{busi}), and X-ray (COVID-QU-Ex \cite{covid}). For \emph{domain generalization}, we test on the breast ultrasound domain (BUID \cite{buid}, BUSBRA \cite{busbra}, UDIAT \cite{udiat}) and the brain MRI domain (BTMRI-P \cite{btmri_p}, BTMRI-S \cite{btmri_s}, BRISC \cite{brisc}). This diverse suite includes challenging modalities, enabling a thorough evaluation across heterogeneous biomedical settings.

\noindent \textbf{Implementation Details.}  
We adopt BiomedCLIP with a ViT-B/16 backbone and report averages over three independent runs. For all experiments, training is performed for 100 epochs using the prompt template ``\texttt{a photo of a [class]}", along with an Adam optimizer \cite{kingma2014adam} with a learning rate of $7.5\times10^{-4}$ and a batch size of 16. All experiments are conducted on one NVIDIA A100 GPU (40GB).

\begin{table*}[t]
\caption{Accuracy (\%) benchmarking for the 9 datasets with BiomedCLIP as the backbone for all methods. \textbf{bold}=best results \& \underline{underlined}=second best results.}
\label{tab:fewshot}
\centering
\resizebox{0.98\textwidth}{!}{
\begin{tabular}{llcccccccccc}
\toprule
Shots & Method & BTMRI & BUSI & COVID & CTKIDNEY & Kvasir & CHMNIST & LC25000 & RETINA & OCTMNIST & Average \\
\midrule
\multirow{1}{*}{0} 
& \textbf{BiomedCLIP} \cite{biomedclip} 
& 56.79 & 59.75 & 43.80 & 42.43 & 54.58 & 30.65 & 50.03 & 26.26 & 30.00 & 43.81 \\
\midrule
\midrule
\multirow{11}{*}{4}
& CoOp \cite{zhou2022learning} 
& 74.68 & 60.17 & 67.03 & 68.12 & 70.78 & 68.66 & 84.66 & 42.22 & 53.37 & 65.52 \\
& CoCoOp \cite{zhou2022conditional} 
& 67.83 & 59.75 & 63.70 & 61.07 & 68.94 & 58.58 & 77.44 & 39.75 & 48.57 & 60.63 \\
& KgCoOp \cite{yao2023visual} 
& 75.40 & 62.01 & 65.91 & \underline{68.68} & 68.28 & 68.77 & 82.10 & 42.61 & 52.97 & 65.19 \\
& ProGrad \cite{zhu2023prompt} 
& 76.24 & 62.29 & 68.56 & 67.90 & 70.00 & 69.13 & 84.72 & 43.09 & 55.07 & 66.33 \\
& BiomedCoOp \cite{koleilat2025biomedcoop} & 77.23 & 59.32 & \textbf{73.28} & 66.50 & \underline{74.08} & 71.19 & \underline{85.60} & 45.58 & 54.73 & \underline{67.50} \\
& LP++ \cite{huang2024lp++} 
& 75.48 & 60.31 & 62.32 & 65.73 & 69.36 & 67.79 & 82.61 & 46.95 & \underline{59.02} & 65.51 \\
& CLIP-Adapter \cite{gao2024clip} 
& 56.80 & 61.72 & 46.28 & 42.19 & 54.83 & 33.26 & 52.91 & 26.07 & 49.96 & 47.11 \\
& TIP-Adapter-F \cite{zhang2021tip} 
& \underline{77.90} & 64.54 & 69.57 & 60.18 & 69.94 & \underline{71.74} & 79.57 & 47.37 & 55.20 & 66.22 \\
& GDA \cite{wang2024hard} 
& 77.56 & 64.41 & 66.77 & 63.41 & 72.19 & 70.92 & 85.00 & \underline{48.42} & 57.42 & 67.34 \\
& CLIP-LoRA \cite{zanella2024low} 
& 77.31 & \underline{64.91} & 69.85 & 66.47 & 66.11 & 64.34 & 84.93 & 47.38 & 52.10 & 65.93 \\
\rowcolor{tabhighlight} & \texttt{Evi-Steer} (Ours) 
& \textbf{78.28} & \textbf{65.54} & \underline{69.93} & \textbf{71.98} & \textbf{74.22} & \textbf{75.29} & \textbf{86.12} & \textbf{49.19} & \textbf{72.33} & \textbf{71.43} \\
\midrule
\midrule
\multirow{11}{*}{8}
& CoOp \cite{zhou2022learning} 
& 79.27 & 64.69 & 74.66 & 77.40 & 77.14 & 75.00 & 87.50 & 51.87 & 63.67 & 72.36 \\
& CoCoOp \cite{zhou2022conditional} 
& 71.69 & 65.82 & 69.36 & 73.93 & 72.92 & 66.58 & 85.57 & 48.45 & 55.40 & 67.75 \\
& KgCoOp \cite{yao2023visual} 
& 79.79 & 67.37 & 74.86 & 77.43 & 72.05 & 69.50 & 84.63 & 49.97 & 61.03 & 70.74 \\
& ProGrad \cite{zhu2023prompt} 
& 78.82 & 64.83 & 74.65 & \underline{78.23} & 76.03 & 70.99 & 87.86 & 52.26 & 62.17 & 71.76 \\
& BiomedCoOp \cite{koleilat2025biomedcoop} & 78.55 & 63.27 & \textbf{76.26} & 77.16 & \underline{77.72} & 74.78 & 88.77 & \underline{56.47} & 58.87 & 72.43 \\
& LP++ \cite{huang2024lp++} 
& 77.11 & 66.10 & 66.19 & 77.06 & 72.52 & 72.40 & 89.14 & 53.44 & 63.69 & 70.85 \\
& CLIP-Adapter \cite{gao2024clip} 
& 57.15 & 61.86 & 48.68 & 44.64 & 56.08 & 36.48 & 56.33 & 25.84 & 49.50 & 48.51 \\
& TIP-Adapter-F \cite{zhang2021tip} 
& 79.18 & 68.50 & 69.89 & 75.24 & 75.86 & 74.51 & \underline{90.31} & 56.07 & 65.00 & 72.73 \\
& GDA \cite{wang2024hard} 
& \underline{81.93} & \underline{70.20} & 74.96 & \textbf{82.33} & 74.47 & 82.03 & 89.17 & 52.58 & \underline{66.57} & \underline{74.92} \\
& CLIP-LoRA \cite{zanella2024low} 
& 78.43 & 69.13 & \underline{75.81} & 77.95 & 70.72 & \underline{82.61} & 86.21 & 50.08 & 61.33 & 72.47 \\
\rowcolor{tabhighlight} & \texttt{Evi-Steer} (Ours) 
& \textbf{83.42} & \textbf{72.74} & 74.69 & 77.71 & \textbf{79.64} & \textbf{82.78} & \textbf{90.39} & \textbf{62.22} & \textbf{72.40} & \textbf{77.33} \\
\midrule
\midrule
\multirow{11}{*}{16}
& CoOp \cite{zhou2022learning} 
& 82.37 & 69.49 & 76.37 & \underline{83.52} & 77.88 & 79.63 & 92.19 & 59.38 & 65.47 & 76.26 \\
& CoCoOp \cite{zhou2022conditional} 
& 78.45 & 70.20 & 74.52 & 77.70 & 75.22 & 72.16 & 87.38 & 53.91 & 60.67 & 72.25 \\
& KgCoOp \cite{yao2023visual} 
& 81.07 & 70.62 & 75.65 & 77.67 & 72.95 & 73.58 & 86.79 & 51.18 & 62.80 & 72.48 \\
& ProGrad \cite{zhu2023prompt} 
& 82.84 & 71.47 & 74.93 & 81.13 & 75.88 & 75.11 & 90.70 & 50.47 & 63.33 & 73.98 \\
& BiomedCoOp \cite{koleilat2025biomedcoop} & \underline{83.30} & 70.34 & \textbf{78.72} & 83.20 & 78.89 & 79.05 & \underline{92.68} & 61.28 & 66.93 & 77.15 \\
& LP++ \cite{huang2024lp++} 
& 81.61 & 70.05 & 72.79 & 79.07 & 75.41 & 78.32 & 92.58 & 60.62 & 68.35 & 75.42 \\
& CLIP-Adapter \cite{gao2024clip} 
& 60.16 & 63.55 & 49.55 & 47.28 & 56.50 & 42.06 & 57.56 & 26.05 & 52.73 & 50.60 \\
& TIP-Adapter-F \cite{zhang2021tip} 
& 82.27 & \underline{71.89} & 76.07 & 82.07 & 78.00 & 80.43 & 92.35 & \underline{62.85} & 72.50 & \underline{77.60} \\
& GDA \cite{wang2024hard} 
& 81.91 & 66.81 & 75.65 & 81.54 & \underline{79.53} & 83.66 & \textbf{93.55} & 57.02 & \underline{75.43} & 77.23 \\
& CLIP-LoRA \cite{zanella2024low} 
& 80.19 & 71.42 & \underline{76.71} & 81.85 & 76.81 & \underline{84.14} & 87.98 & 52.19 & 61.50 & 74.75 \\
\rowcolor{tabhighlight} & \texttt{Evi-Steer} (Ours) 
& \textbf{86.39} & \textbf{73.16} & 73.44 & \textbf{90.11} & \textbf{82.33} & \textbf{84.22} & 92.05 & \textbf{68.53} & \textbf{80.37} & \textbf{81.18} \\
\bottomrule
\end{tabular}}
\end{table*}

\begin{table*}[ht]
\centering
\caption{Domain generalization accuracy (\%):
models are trained on a source dataset with 16-shots (except BiomedCLIP) and then directly evaluated on OOD target datasets without adaptation.}
\setlength{\tabcolsep}{3pt}
\resizebox{0.77\linewidth}{!}{%
\begin{tabular}{l cccc cccc}
\toprule
{\multirow{4}{*}{\textbf{Method}}} & \multicolumn{4}{c}{\textbf{Breast Ultrasound}} 
& \multicolumn{4}{c}{\textbf{Brain MRI}} \\
\cmidrule(lr){2-5} \cmidrule(lr){6-9}
& \textbf{Source} & \multicolumn{3}{c}{\textbf{Target}} 
& \textbf{Source} & \multicolumn{3}{c}{\textbf{Target}} \\
\cmidrule(lr){2-2} \cmidrule(lr){3-5} \cmidrule(lr){6-6} \cmidrule(lr){7-9} 
& BUSI & BUID & BUSBRA & UDIAT 
& BTMRI & BTMRI-P & BTMRI-S & BRISC \\
\midrule
\textcolor{mutedblue}{BiomedCLIP} \cite{biomedclip} & $59.75$ & $75.00$ & $66.78$ & $61.54$ & $56.79$ & $52.80$ & $55.20$ & $52.60$ \\
CoOp \cite{zhou2022learning} & $69.49$ & $72.22$ & $68.28$ & $70.49$ & $82.37$ & $76.77$ & $78.13$ & $74.87$ \\
CoCoOp \cite{zhou2022conditional} & $70.20$ & $67.59$ & $66.55$ & $71.54$ & $78.45$ & $75.43$ & $76.62$ & $70.20$ \\
ProGrad \cite{zhu2023prompt} & $\underline{71.47}$ & $69.44$ & $67.25$ & $69.23$ & $82.63$ & $\underline{78.00}$ & $\underline{79.96}$ & $\underline{76.27}$ \\
KgCoOp \cite{yao2023visual} & $70.62$ & $\underline{77.78}$ & $\underline{68.79}$ & $\underline{75.64}$ & $81.07$ & $75.60$ & $79.02$ & $73.37$ \\
GDA \cite{wang2024hard} & $66.81$ & $63.89$ & $63.02$ & $67.95$ & $81.91$ & $77.47$ & $76.75$ & $75.23$ \\
CLIP-LoRA \cite{zanella2024low} & $71.42$ & $65.85$ & $65.37$ & $69.82$ & $80.19$ & $77.90$ & $77.68$ & $73.50$ \\
BiomedCoOp \cite{koleilat2025biomedcoop} 
& $70.34$ & $67.59$ & $62.90$ & $71.67$ 
& $\underline{83.30}$ & $77.60$ & $79.52$ & $74.50$ \\
\midrule
\rowcolor{tabhighlight} \texttt{Evi-Steer} (Ours)
& $\mathbf{73.16}$ & $\mathbf{78.70}$ & $\mathbf{71.61}$ & $\mathbf{78.21}$ 
& $\mathbf{86.39}$ & $\mathbf{80.23}$ & $\mathbf{80.53}$ & $\mathbf{78.43}$ \\
\bottomrule
\end{tabular}}
\label{tab:domain_generalization}
\end{table*}

\section{Results}
\subsection{Few-shot Adaptation \& Domain Generalization Evaluation}
\noindent \textbf{Few-shot adaptation.} As shown in Table~\ref{tab:fewshot}, \texttt{Evi-Steer} consistently achieves the best overall performance across all $K$-shot regimes, improving the average accuracy from $43.81\%$ (zero-shot) to $71.45\%$, $77.34\%$, and $81.21\%$ at $K=4/8/16$ shots, with strong gains on challenging datasets (e.g. BTMRI, OCTMNIST), updating only $0.11\%$ of the total parameters (221K parameters), on par with CLIP-LoRA in parameter budget, while consistently surpassing it in performance.

\noindent \textbf{Domain generalization.} Table~\ref{tab:domain_generalization} shows that \texttt{Evi-Steer} generalizes better under domain shift, yielding the strongest OOD transfer in both breast Ultrasound and brain MRI domains (e.g., BUSI$\rightarrow$BUID: $78.70\%$, BTMRI$\rightarrow$BTMRI-P: $80.23\%$). Note the higher in-distribution (ID) performance compared to OOD in the breast ultrasound domain is expected, as the OOD dataset lacks the \emph{normal scan} class, whereas BUSI contains three classes (malignant/benign/normal).

\begin{figure}[t]
\centering
\includegraphics[width=0.75\linewidth, height=0.3\linewidth]{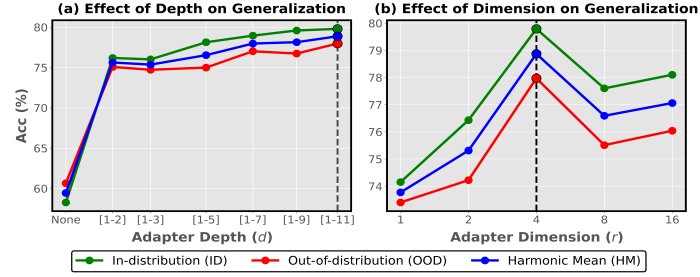}
\caption{Effect of layer interventions and adapter dimension on accuracy.}
\label{fig:layers_dimension}
\end{figure}

\begin{table}[!htbp]
\caption{Effect of different key components in \texttt{Evi-Steer} (HM=harmonic mean).}
\tablestyle{-22pt}{1.0}
\addtolength{\tabcolsep}{+24pt}
\resizebox{0.73\linewidth}{!}{
\centering
\begin{tabular}{lccc}
\toprule
\multicolumn{1}{c}{\multirow{2}{*}{\textbf{Method}}} & 
\multicolumn{3}{c}{\textbf{Domain Generalization}} \\ 
\cmidrule(lr){2-4}
 & \textbf{ID Acc. (\%)} & \textbf{OOD Acc. (\%)} & \textbf{HM Acc. (\%)} \\
\midrule
\rowcolor{tabhighlight} \texttt{Evi-Steer} (Ours) & \textbf{79.79} & \textbf{77.97} & \textbf{78.87} \\
\midrule
w/o Visual Adaptation & 76.23$_{(-3.56)}$\textcolor{red}{$\downarrow$} & 72.25$_{(-5.72)}$\textcolor{red}{$\downarrow$} & 74.19$_{(-4.68)}$\textcolor{red}{$\downarrow$} \\
w/o Textual Adaptation & 78.40$_{(-1.39)}$\textcolor{red}{$\downarrow$} & 76.97$_{(-1.00)}$\textcolor{red}{$\downarrow$} & 77.68$_{(-1.19)}$\textcolor{red}{$\downarrow$} \\
w/o Evidential Update & 79.25$_{(-0.54)}$\textcolor{red}{$\downarrow$} & 76.34$_{(-1.63)}$\textcolor{red}{$\downarrow$} & 77.77$_{(-1.10)}$\textcolor{red}{$\downarrow$} \\
w/o Cross-modal Belief & 79.52$_{(-0.27)}$\textcolor{red}{$\downarrow$} & 76.90$_{(-1.07)}$\textcolor{red}{$\downarrow$} & 78.19$_{(-0.68)}$\textcolor{red}{$\downarrow$} \\
\bottomrule
\end{tabular}
}
\label{tab:component_ablation}
\end{table}

\subsection{Ablation Experiments}
\noindent \textbf{Effectiveness of components.}
Table \ref{tab:component_ablation} shows that all components contribute to domain generalization, with visual adaptation being most critical. Removing it causes the largest degradation ($-5.72\%$ OOD; $-4.68\%$ harmonic mean (HM)), highlighting the need to adapt visual representations. Textual adaptation also consistently contributes, yielding $-1.00\%$ OOD and $-1.19\%$ HM drops. The evidential update and cross-modal belief modules mainly affect OOD performance, suggesting uncertainty-aware learning improves robustness under domain shift.

\noindent \textbf{Effect of adapter depth.}
Figure~\hyperref[fig:layers_dimension]{\ref*{fig:layers_dimension}a} shows that increasing the number of adapted layers $d$ consistently improves generalization. Performance rises from no adaptation (HM $59.44\%$) to shallow adaptation and continues improving with depth, reaching the best performance at depth 11 (HM $78.87\%$). This monotonic trend indicates that distributing lightweight adapters across the network is more effective than shallow adaptation, enabling progressive cross-modal refinement.

\noindent \textbf{Effect of adapter dimension.}
We evaluate the adapter dimension $r \in \{1,2,4,\allowbreak 8, 16\}$, as shown in Fig.~\hyperref[fig:layers_dimension]{\ref*{fig:layers_dimension}b}. Performance improves from $r=1$ to $r=4$, reaching a peak at $r=4$ with the best HM accuracy of $78.87\%$. Larger dimensions provide no further gains and slightly degrade performance, suggesting that moderate-dimensional adapters best balance capacity and overfitting, and that \texttt{Evi-Steer} benefits from dimension-constrained updates.

\section{Conclusion}
We introduced \texttt{Evi-Steer}, a novel parameter-efficient adaptation framework for biomedical VLMs, by combining low-dimensional cross-modal updates with evidential uncertainty modeling and confidence-based gating. With strong few-shot and domain-generalization results, \texttt{Evi-Steer} demonstrates the value of confidence-aware biomedical vision-language adaptation in clinical settings. Future work will extend the approach to dense prediction and interactive workflows.

\begin{credits}
    \subsubsection{\ackname} We acknowledge the support of the Natural Sciences and Engineering Research Council of Canada and the Fonds de recherche du Qu\'ebec – Nature et technologies (B2X-363874).
\end{credits}

\bibliographystyle{ieeenat_fullname}
\bibliography{main}

\end{document}